%% file: main.tex
\documentclass[10pt]{article} % For LaTeX2e
\usepackage[preprint]{tmlr}
\input{math_commands.tex}

\usepackage{hyperref}
\usepackage{url}

\usepackage{amsthm}
\usepackage{amssymb}

\usepackage{tikz}
\usetikzlibrary{calc}
\usepackage{cleveref}
\usepackage{listings}
\usepackage{pifont}
\usepackage{booktabs}

\usepackage{algorithm}
\usepackage{algpseudocode} 
\usepackage{algorithmicx}

\newtheorem{definition}{Definition}

\algnewcommand{\FindValidSets}{\textsc{FindValidAdjustmentSets}}
\algnewcommand{\FindCritEdges}{\textsc{FindCriticalEdges}}
\algnewcommand{\CheckValidBD}{\textsc{CheckBackdoorValidity}}

\hypersetup{
    colorlinks = true,
    citecolor  = blue,
    linkcolor  = blue
}
\title{\emph{Technical Report:} Facilitating the Adoption of Causal Inference Methods Through LLM-Empowered Co-Pilot}

\author{\name Jeroen Berrevoets\thanks{Work done while at University of Cambridge} \email jeroen.berrevoets@ataraxis.ai \\
      Ataraxis.ai
      \AND
      \name Julianna Piskorz \email jp2048@cam.ac.uk \\
      \addr University of Cambridge
      \AND
      \name Rob Davis \email rd473@cam.ac.uk\\
      \addr University of Cambridge
      \AND
      \name Harry Amad \email hmka3@cam.ac.uk \\
      \addr University of Cambridge 
      \AND
      \name Jim Weatherall \email \\
      \addr AstraZeneca 
      \AND
      \name Mihaela van der Schaar \email mv472@cam.ac.uk \\
      \addr University of Cambridge
      }

\def\month{MM}  % Insert correct month for camera-ready version
\def\year{YYYY} % Insert correct year for camera-ready version
\def\openreview{\url{https://openreview.net/forum?id=XXXX}} % Insert correct link to OpenReview for camera-ready version

\begin{document}

\maketitle

\begin{abstract}
Estimating treatment effects (TE) from observational data is a critical yet complex task in many fields, from healthcare and economics to public policy. While recent advances in machine learning and causal inference have produced powerful estimation techniques, their adoption remains limited due to the need for deep expertise in causal assumptions, adjustment strategies, and model selection. In this paper, we introduce \texttt{CATE-B}, an open-source co-pilot system that uses large language models (LLMs) within an agentic framework to guide users through the end-to-end process of treatment effect estimation. \texttt{CATE-B} assists in (i) constructing a structural causal model via causal discovery and LLM-based edge orientation, (ii) identifying robust adjustment sets through a novel Minimal Uncertainty Adjustment Set criterion, and (iii) selecting appropriate regression methods tailored to the causal structure and dataset characteristics. To encourage reproducibility and evaluation, we release a suite of benchmark tasks spanning diverse domains and causal complexities. By combining causal inference with intelligent, interactive assistance, \texttt{CATE-B} lowers the barrier to rigorous causal analysis and lays the foundation for a new class of benchmarks in automated treatment effect estimation.
\end{abstract}

\section{Introduction}

% who is this meant for
% with the agentic framework, we are bringing a wealth of machine learning tools developed over the previous 10 years to people who can actually use it
% everybody can interact with the tool in the same way

\paragraph{Treatment Effect Inference Offers a Powerful Tool for Data-Driven Decision Making.}
Estimating treatment effects (TE) is pivotal in guiding decisions across various domains by quantifying the causal impact of a treatment $W$ on an outcome $Y$ within a target population. For instance, in medicine, $W$ could represent a new drug and $Y$ the health outcome; in marketing, $W$ might be an advertising campaign and $Y$ the resulting sales; and in policy design, $W$ could be a legislative change with $Y$ reflecting its societal impact. The most rigorous method for estimating TE is through randomized controlled trials (RCTs). However, RCTs are often impractical due to ethical considerations, logistical challenges, or high costs.

In such scenarios, inferring treatment effects from observational data becomes essential. Advanced statistical and machine learning techniques have been developed to estimate causal effects despite the presence of potential confounding biases. Over the years, significant advances in identification and adjustment techniques as well as regression methods \citep{kunzel_metalearners_2019, shi_adapting_2019, kennedy_towards_2023} have improved the reliability of causal estimates, facilitating causal analysis in the absence of experimental data.

\paragraph{Challenges in the Widespread Adoption of Treatment Effect Inference Methods.}
Despite the substantial benefits of treatment effect (TE) inference techniques, their widespread adoption remains limited by several challenges. These methods often require deep expertise to correctly specify causal assumptions and implement appropriate estimation strategies. Unlike predictive models, causal inference methods cannot be reliably validated using held-out datasets; instead, validity hinges on untestable assumptions and domain-specific knowledge, demanding costly collaborations between statistical and subject-matter experts. Although recently open-source software libraries \citep{econml} have facilitated access to treatment effect estimation methods with minimal code, thus simplifying implementation, these frameworks often assume that the adjustment set simply consists of all observed covariates -- a naïve assumption that frequently leads to biased estimates in complex, real-world scenarios \citep{pearl2009myth, pearl2010class, elwert2014endogenous}.

While structural causal models (SCMs) and do-calculus provide a principled basis for deriving minimal valid adjustment sets \citep{pearl2009causal, smucler2021}, their practical application remains difficult. Constructing an SCM requires accurately specifying a full causal graph, a task that is often infeasible without extensive domain expertise. Moreover, deriving valid adjustment sets involves complex graphical criteria, and selecting appropriate regression methods for estimation introduces further challenges. As a result, practitioners often avoid rigorous causal frameworks, instead defaulting to simpler -- but potentially biased -- approaches.

\paragraph{The Need for Intelligent, Accessible Causal Inference Tools.} There is a pressing need for intelligent systems that can guide users through the intricacies of causal analysis. Recent advances in agentic machine learning frameworks have demonstrated that large language models (LLMs) can be used not just for text generation, but also for structured problem solving and planning in complex tasks \citep{yao2023react, shinn2023reflexion}. By embedding reasoning, tool use, and iterative feedback within LLM agents, it becomes possible to empower non-expert users to carry out tasks that traditionally required domain expertise. We argue that treatment effect estimation is an ideal candidate for such a deployment: a system that assists users through graph construction, adjustment set derivation, and regression modelling can dramatically lower the barrier to rigorous causal analysis.

\begin{figure}
    \centering
    \includegraphics[width=\linewidth]{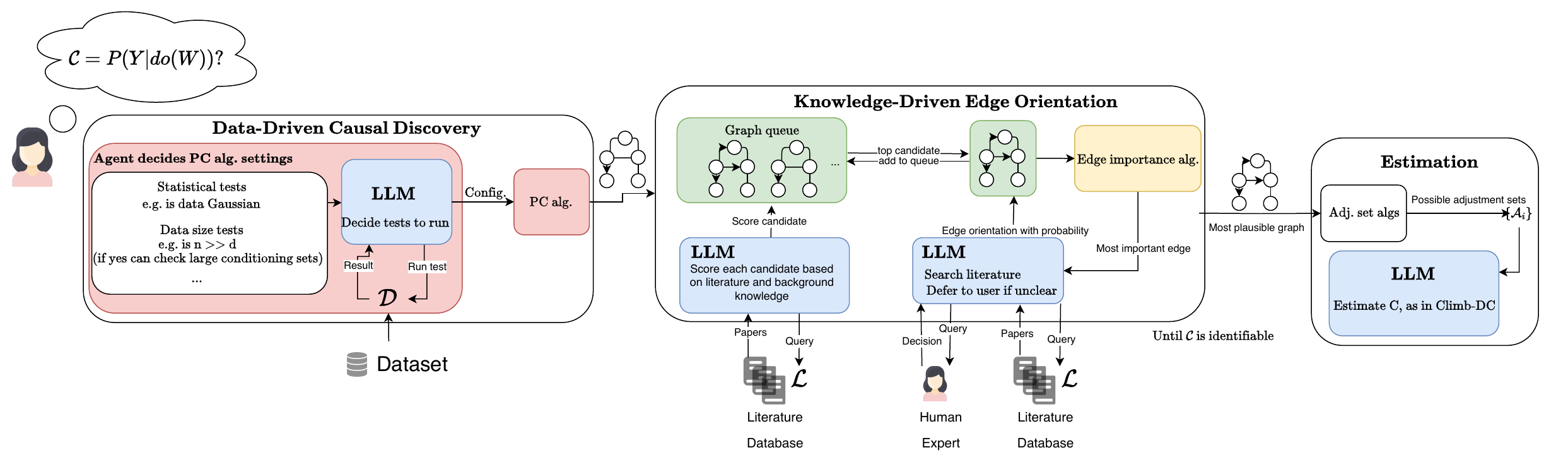}
    \caption{Overview of the proposed LLM Co-Pilot.}
    \label{fig:fig1}
\end{figure}

\paragraph{Enhancing the Adoption of Treatment Effect Inference Through an LLM-Guided Co-Pilot.}
To address these challenges, we propose a novel open-source co-pilot system, \texttt{CATE-B},\footnote{Named after the actress Cate Blanchett, whose work and talent inspire us on a daily basis.} that uses large language models (LLMs) within an agentic framework to facilitate rigorous treatment effect analysis. This system offers a user-friendly interface, enabling practitioners to perform efficient causal analyses without extensive coding or deep expertise in causal inference methodologies. Our framework is visualized in \cref{fig:fig1}.

The contributions of our work can be summarised as follows:
\begin{itemize}
    \item \textit{Construction of a Complete Graphical Model:}
    The co-pilot assists users in constructing a comprehensive SCM by combining classical causal discovery algorithms—such as PC, GES, and FCI—with advanced LLM capabilities. These algorithms identify a Markov equivalence class of directed acyclic graphs (DAGs), which the co-pilot refines by querying external academic resources to orient ambiguous edges, thereby producing a precise causal graph.

    \item \textit{Identification of Appropriate Adjustment Sets:}  
    Leveraging the constructed DAG, the co-pilot translates the practitioner's causal query into suitable adjustment sets. To improve robustness, we introduce the concept of a \textit{Minimal Uncertainty Adjustment Set}, which accounts for LLM-derived uncertainties about edge orientation to derive adjustment sets that are valid across plausible causal structures.

    \item \textit{Selection of Robust Regression Techniques:}  
    The co-pilot then guides users in selecting appropriate regression techniques. It incorporates user preferences, dataset metadata, and knowledge from similar prior studies to recommend estimation strategies that align with the causal structure and data characteristics.
\end{itemize}

By integrating these functionalities, \texttt{CATE-B} empowers researchers, data scientists, clinicians, and policy analysts -- our primary target audience -- to conduct rigorous and efficient causal analyses. Furthermore, the open-source nature of \texttt{CATE-B} ensures transparency, extensibility, and wide accessibility, aiming to democratize advanced causal inference capabilities. By enabling structured implementation of causal inference pipelines across a variety of datasets and decision scenarios, we provide a foundation for future benchmarking efforts in automated causal analysis.

% explain that neither causal inference nor practitioner people can solve SCMs on their own -- which is why there is a need for the co-pilot

\section{Background and Problem Formalism}

The main strategy for inferring ATE is adjustment \citep{imbens2015causal}, where every non-causal association is to be removed from our causal estimate.
An example adjustment is the backdoor adjustment. 
Backdoor adjustment assumes a fairly standard graphical causal structure, 
\begin{tikzpicture}[
    baseline=-1mm
]
    \node[inner sep=0] (x) at (0,0) {$X$};
    \node[inner sep=0] (y) at ($(x) + (1, 0)$) {$Y(w)$};
    \node[inner sep=0] (w) at ($(x) - (.8, 0)$) {$W$};

    \draw[-latex] (x) -- (y);
    \draw[-latex] (x) -- (w);
\end{tikzpicture},
which clearly shows that $X$ is confounding our treatment, $W$, and potential outcomes $Y(w)$.
Hence, we need to adjust our causal estimate, $\mathbb{E}[Y(1) - Y(0)]$, through $X$:
\begin{equation} \label{eq:backdoor}
    \mathbb{E}[Y(1) - Y(0)] = \mathbb{E}_X\big[\mathbb{E}[Y|W=1, X] - \mathbb{E}[Y|W=0, X]\big].
\end{equation}
Where the rhs in \cref{eq:backdoor}, no longer contains any causal objects such as $Y(w)$.

\subsection{Assumptions Underlying Treatment Effect Estimation}

Estimating treatment effects (TE) from observational data hinges on a set of core assumptions that permit causal identification. While the backdoor adjustment (in \cref{eq:backdoor}) formula--- derived from graphical models \citep{pearl2009causal} ---offers a principled approach to isolating the effect of a treatment $W$ on an outcome $Y$, its validity depends on several subtle and often underappreciated conditions.

First, we require the no {\bf interference assumption}, often formalized as part of the Stable Unit Treatment Value Assumption (SUTVA) \citep{imbens2015causal}. This stipulates that the treatment assignment of one unit does not affect the outcome of another. Though often taken for granted, this assumption is easily violated in real-world settings with spillover or network effects--- such as vaccination programs, where an individual’s likelihood of infection may be influenced by the vaccination status of their peers. In such cases, the observed outcomes cannot be cleanly attributed to individual-level treatment assignments, undermining the causal estimand.

Another foundational assumption is {\bf (strong) ignorability} of the treatment assignment. Here, we assume that conditional on a set of observed covariates $X$, treatment assignment is as good as random--- formally, $Y(w) \perp W |X$. This implies that all confounding pathways between treatment and outcome are blocked by $X$, allowing for unbiased adjustment. While various conditional independence tests exist, ignorability is fundamentally untestable because it pertains to counterfactuals—quantities we can never observe directly. Moreover, this assumption is frequently violated in practice due to unobserved confounders—latent variables that influence both $W$ and $Y$ but are missing from the dataset. In the presence of such unobserved structure, estimates of TE may be severely biased.

The assumption of {\bf positivity}, or overlap, is also essential. It requires that every individual in the population has a non-zero probability of receiving each treatment level, given their covariates. That is, for all values of $X$, we must have $0<p(W=w | X=x)<1$. Violations occur when certain subgroups are deterministically assigned to a treatment or control condition—often the case in observational data when ethical or logistical constraints prevent treatment in some subpopulations. In high-dimensional or continuous covariate spaces, ensuring sufficient overlap becomes especially challenging, and positivity is practically unverifiable in finite samples.

Finally, the assumption of consistency connects observed data to potential outcomes. It requires that if an individual receives treatment $W=w$, then their observed outcome $Y$ is equal to the potential outcome $Y(w)$. This implies that the treatment is well-defined and uniformly administered; that is, there are no multiple versions of treatment, and no ambiguity in its implementation or measurement. Inconsistent treatment delivery, non-adherence, or misclassification can all compromise this assumption.

While these assumptions are sufficient to identify treatment effects nonparametrically, practical estimation typically introduces additional parametric assumptions. Popular CATE estimation methods, such as those implemented in libraries like \texttt{DoWhy}, \texttt{EconML}, and \texttt{CausalML}, adopt meta-learner strategies (e.g., T-, S-, X-learners) that reduce causal inference to supervised learning tasks \citep{kunzel_metalearners_2019}. These approaches often rely on black-box regressors—ranging from linear models to neural networks—which impose their own assumptions, such as linearity, smoothness, or differentiability. While flexible, such estimators may perform poorly if these implicit modeling assumptions are violated. Moreover, their performance is often sensitive to hyperparameter choices, sample size, and the underlying data distribution.

Some of these parametric assumptions can be diagnosed via statistical tools—such as residual diagnostics, goodness-of-fit tests, or model comparison metrics--- but doing so typically requires statistical expertise that many domain practitioners may lack. Thus, even when the core identification assumptions hold, mismatched model choices can lead to inaccurate estimates or misleading policy conclusions.

\subsection{The Need for an Intelligent Causal Co-Pilot}
The layered complexity of causal inference—from untestable assumptions to model specification—creates a substantial barrier to the widespread and reliable application of treatment effect estimation methods. Effective TE estimation demands a synthesis of causal expertise, to validate the identification strategy, and domain knowledge, to assess the plausibility of assumptions in context. In practice, coordinating these areas of expertise is expensive, time-consuming, and often infeasible—particularly in high-stakes or fast-moving decision environments such as public health, marketing, or policy analytics.

To address this gap, we propose \texttt{CATE-B}, an intelligent co-pilot system powered by large language models. \texttt{CATE-B} is designed to assist users through the full causal inference pipeline: constructing a plausible causal graph, deriving robust adjustment sets, and selecting appropriate regression strategies tailored to the data at hand. By embedding structured reasoning, domain adaptation, and tool-use within an LLM framework, \texttt{CATE-B} reduces the need for manual intervention and lowers the barrier to entry for rigorous causal analysis.

In \cref{sec: cate-b}, we detail the system’s architecture and capabilities, and in \cref{sec:experiments}, we present empirical results demonstrating how \texttt{CATE-B} facilitates treatment effect estimation across a range of benchmark datasets and domains.

\section{CATE-B: LLM-Empowered Co-Pilot for Treatment Effect Estimation}
\label{sec: cate-b}

Our co-pilot works in three phases:
\begin{enumerate}
    \item \textbf{Phase 1:} Using a user-provided dataset, we infer a Markov equivalence class of the underlying structural causal models (SCMs). Then, with the help of the LLM we query scientific evidence on how each two variables relate with each other causally, which allows us to orient the edges in the SCM.
    \item \textbf{Phase 2:} We then proceed to translate the identified SCM into a minimal adjustment set, guiding the identification of the adjustment set by the uncertainty in the edge orientation. To do that, we formalise the notion of a \textit{Minimal Uncertainty Adjustment Set}.
    \item \textbf{Phase 3:} After obtaining the adjustment set $Z \in \mathcal{Z} \subseteq \mathcal{X}$, we then proceed to regress the outcomes and obtain the estimate of the treatment effect using simple adjustment: \begin{equation}
    \mathbb{E}_Z[\mathbb{E}[Y|W=1, Z] - \mathbb{E}[Y|W=0, Z]].
    \end{equation}
    Working with $Z$, derived from a SCM, we allow much more complicated settings than what is possible with standard CATE techniques, yet still allow for ease-of-use.
\end{enumerate}

We describe each phase of the co-pilot in more detail below.

\subsection{Phase 1: Recovering the SCM, with RAG-empowered edge orientation}

The first phase of our co-pilot constructs a causal graph over the variables in a user-provided dataset, with the goal of approximating a valid Structural Causal Model (SCM). This begins by applying standard structure learning algorithms-- such as the PC or FCI algorithm --to recover a Markov equivalence class of directed acyclic graphs (DAGs) that are observationally indistinguishable from the data alone. These graphs encode the conditional independence structure but leave some edges unoriented due to limitations of purely statistical methods.

To resolve edge orientation within this equivalence class, we introduce a retrieval-augmented generation (RAG) pipeline that queries scientific literature to identify domain-specific causal relationships. For each unoriented or contested edge $X \leftrightarrow Y$, the system constructs a query reflecting the causal question (e.g., “Does $X$ cause $Y$ in human biology?”) and retrieves relevant scientific abstracts or evidence from curated sources. These are passed to a large language model (LLM), which synthesizes the information and provides a proposed orientation, along with a natural language explanation and a confidence score $c(X\rightarrow Y)\in[0,1]$.

The result is a partially oriented DAG $G=(V,E)$, where a subset of edges $\mathcal{E}_{\textrm{oriented}}\subseteq E$ are assigned directionality by the LLM-RAG pipeline, while others remain uncertain or undirected. Importantly, the associated uncertainty scores $u(e)=1-c(e)$ are preserved and propagated to downstream components, enabling principled reasoning about structural ambiguity in subsequent phases. This fusion of statistical discovery and external knowledge retrieval equips the co-pilot with a hybrid epistemic foundation—grounding edge orientation not only in data-driven patterns but also in contextual scientific understanding.

\subsection{Phase 2: Finding the Minimum Uncertainty Adjustment Set}

Phase 1 of our co-pilot yields an inferred causal graph $\mathcal{G}=(V, E)$ where the orientation of certain edges $\mathcal{E}_{oriented} \subseteq E$ carries uncertainty, quantified by confidence scores $c(e)$ derived from LLM-human collaboration. Phase 2 aims to find an adjustment set $Z$ to identify the causal effect of interest, $P(Y|do(W))$. Standard graphical criteria, like Pearl's back-door criterion, identify multiple \textit{valid} adjustment sets $\mathcal{Z}_{valid}$. Choosing among these often involves heuristics like minimizing cardinality or optimizing for statistical efficiency.

However, these standard selection methods do not leverage the uncertainty $u(e) = 1 - c(e)$ associated with the graph structure itself. Relying on an adjustment set whose validity hinges on low-confidence edge orientations introduces fragility into the causal estimate. To address this, we propose selecting the adjustment set that minimizes the reliance on uncertain structural assumptions. We formalize this through the Minimum Uncertainty Adjustment Set (MUAS).

\begin{definition}[Minimum Uncertainty Adjustment Set (MUAS)]
Let $\mathcal{G}=(V, E)$ be a DAG with treatment $W \in V$ and outcome $Y \in V$. Let $\mathcal{E}_{oriented} \subseteq E$ be a subset of edges with associated uncertainty $u: \mathcal{E}_{oriented} \to [0, 1]$. Define $u(e) = 0$ for $e \in E \setminus \mathcal{E}_{oriented}$.
Let $\mathcal{C}$ be a criterion for identifying $P(Y|do(W))$ via adjustment (e.g., back-door), and let $\mathcal{Z}_{valid}(\mathcal{G}, W, Y, \mathcal{C})$ be the set of all valid adjustment sets $Z \subseteq V \setminus \{W, Y\}$ according to $\mathcal{C}$ in $\mathcal{G}$.

For $Z \in \mathcal{Z}_{valid}$ and $e \in \mathcal{E}_{oriented}$, let $\mathcal{G}'_e$ be the graph $\mathcal{G}$ with the orientation of $e$ reversed. The set of \textbf{critical edges} for $Z$ is $\mathcal{E}_{critical}(Z) = \{e \in \mathcal{E}_{oriented} \mid Z \notin \mathcal{Z}_{valid}(\mathcal{G}'_e, W, Y, \mathcal{C})\}$.
\\

The \textbf{Uncertainty Cost} of $Z \in \mathcal{Z}_{valid}$ is $\text{Cost}_U(Z) = \max(\{u(e) \mid e \in \mathcal{E}_{critical}(Z)\})$.

A \textbf{Minimum Uncertainty Adjustment Set (MUAS)} $Z^*$ is a set such that: $Z^* \in \underset{Z \in \mathcal{Z}_{valid}(\mathcal{G}, W, Y, \mathcal{C})}{\operatorname{argmin}} \text{Cost}_U(Z).$
\end{definition}

The MUAS can be identified algorithmically as follows:

\begin{algorithm}[H]
\caption{Find Minimum Uncertainty Adjustment Set (MUAS)}
\label{alg:find_muas_concise}
\begin{algorithmic}[1]
\State \textbf{Input:} DAG $\mathcal{G}=(V, E)$, Treatment $W$, Outcome $Y$, Uncertainties $u(e)$ for $e \in E$, Criterion $\mathcal{C}$.
\State \textbf{Output:} A Minimum Uncertainty Adjustment Set $Z_{muas}$.

\State $\mathcal{S}_{cand} \gets \FindValidSets(\mathcal{G}, W, Y, \mathcal{C})$ 
    \Comment{Find candidate adj. sets}
\If{$\mathcal{S}_{cand} = \emptyset$} \Return Failure \EndIf

\State $min\_cost \gets \infty$
\State $Z_{muas} \gets \text{null}$

\For{each candidate set $Z \in \mathcal{S}_{cand}$}
    \State $\mathcal{E}_{crit} \gets \FindCritEdges(Z, \mathcal{G}, W, Y, \mathcal{C})$ 
        \Comment{Identify essential edges for $Z$'s validity}
    \State $current\_cost \gets 0$
    \If{$\mathcal{E}_{crit} \neq \emptyset$}
        \State $current\_cost \gets \max_{e \in \mathcal{E}_{crit}} \{u(e)\}$
    \EndIf
    
    \If{$current\_cost < min\_cost$}
        \State $min\_cost \gets current\_cost$
        \State $Z_{muas} \gets Z$
    \EndIf
\EndFor

\State \Return $Z_{muas}$
\end{algorithmic}
\end{algorithm}

\begin{algorithm}[H]
\caption{Find Critical Edges (Sensitivity Analysis for Back-door)}
\label{alg:find_critical_edges}
\begin{algorithmic}[1]
\State \textbf{Input:} DAG $\mathcal{G}=(V, E)$, Treatment $W$, Outcome $Y$, Candidate Set $Z$, Set of oriented edges $\mathcal{E}_{oriented} \subseteq E$.
\State \textbf{Output:} Set of critical edges $\mathcal{E}_{critical}$.

\State $\mathcal{E}_{critical} \gets \emptyset$ \Comment{Step 1: Initialize}

\For{each edge $e = (u \to v) \in \mathcal{E}_{oriented}$} \Comment{Step 2: Test edge sensitivity}
    \State $\mathcal{G}' \gets \mathcal{G}$ with edge $e$ replaced by $e' = (v \to u)$.    
    \If{not $\CheckValidBD(\mathcal{G}', W, Y, Z)$}
        \State $\mathcal{E}_{critical} \gets \mathcal{E}_{critical} \cup \{e\}$ 
            \Comment{Original orientation of e is essential}
    \EndIf
\EndFor

\State \Comment{Step 4: Return critical edges}
\State \Return $\mathcal{E}_{critical}$
\end{algorithmic}
\end{algorithm}

\paragraph{Robustness through MUAS.}
By minimizing the maximum uncertainty among critical edges, the MUAS criterion selects an adjustment strategy that avoids relying heavily on any single, highly uncertain structural assumption, prioritizing epistemic robustness. An MUAS might be larger than a minimal cardinality set if the latter's validity depends critically on a low-confidence edge orientation. Using the MUAS therefore yields an adjustment procedure, and consequently a causal effect estimate (Phase 3), that is more robust and less sensitive to the specific uncertainties arising from the LLM-driven graph discovery in Phase 1. This aligns with the co-pilot's goal of providing reliable causal inference from observational data even when the underlying causal structure is not perfectly known.

\subsection{Phase 3: Obtaining the Treatment Effect}

With a robust adjustment set $Z\subseteq X$, hand—selected through the MUAS criterion in Phase 2-- we proceed to estimate the Conditional Average Treatment Effect (CATE). Under standard identification assumptions (e.g., consistency, positivity, and conditional ignorability given $Z$), the treatment effect is computed via covariate adjustment using the following estimand:
\begin{equation}\label{eq:cate:adjustment}
    \mathbb{E}_Z[\mathbb{E}[Y | X, Z, W=1] - \mathbb{E}[Y | X, Z, W=0]].
\end{equation}
This approach re-weights outcome predictions based on the observed covariate distribution to approximate the counterfactual difference in outcomes had the treatment been assigned uniformly.

Importantly, the use of a SCM-derived adjustment set $Z$ enables valid estimation even in complex settings—such as those involving collider bias or latent confounding pathways—that would be difficult to address using default variable selections or off-the-shelf regression tools. Furthermore, our co-pilot integrates with standard supervised learning frameworks (e.g., \texttt{scikit-learn}, \texttt{EconML}) to fit the conditional expectations 
$\mathbb{E}[E | X, Z, W]$ using any compatible model, including linear regressors, random forests, or neural networks. This ensures both flexibility in estimator choice and ease of integration into existing workflows.

By explicitly grounding the adjustment set in a causal graph-- rather than relying on heuristic feature selection --the final treatment effect estimate is more interpretable, statistically justified, and robust to common structural misspecifications. This completes the end-to-end causal reasoning pipeline, turning raw observational data into actionable causal insights with the guidance of \texttt{CATE-B}.

\section{The CATE-B interface}
\subsection{Motivation and Scope}
Practitioners seeking to answer interventional questions from observational data face a steep learning curve: they must stitch together disparate libraries for graph discovery, do-calculus adjustment, and treatment-effect regression, then write custom glue code to orchestrate them. On top of that, a proper analysis demands both deep causal-inference expertise and intimate knowledge of the domain data-yet machine-learning engineers often lack the nuances of the application setting, while subject-matter experts are unfamiliar with the intricacies of causal methods. This separation too frequently results in hidden, unvalidated assumptions and ultimately biased or inaccurate effect estimates.

To address these hurdles, we introduce CATE-B, a fully no-code, chatbot-driven platform that unifies every step of the causal-inference workflow under one conversational interface. Behind the scenes, CATE-B offers a plug-in framework for DAG discovery (e.g.\ PC, GES, NO-TEARS), multiple ATE estimators (e.g.\ TARNET, CFRNET, BART), and an LLM-powered edge-orientation layer that queries scientific literature to resolve ambiguities. Users simply upload their data, pose a natural-language query (e.g. “Does treatment X reduce outcome Y?”), and the system automatically recovers a structural causal model, derives a minimal adjustment set, fits an appropriate regression estimator, and returns both point estimates and diagnostic reports-entirely via the conversational interface, without manual scripting.

% \subsection{High-Level Architecture}
% \begin{figure}
%     \centering
%     \includegraphics[width=0.5\linewidth]{Causal_copilot_architecture_wide.png}
%     \caption{The high-level architecture of CATE-B showing 1) how the benchmarking environment allows for CATE estimation either after or before LLM-empowered causal discovery and adjustment set calculation and 2) how the modular, plug-in architecture allows for complete inter-operability of different methods.}
%     \label{fig:architecture-diagram}
% \end{figure}

\subsection{Core functions of the treatment effects estimation}
\begin{figure}
    \centering
    \includegraphics[width=.7\linewidth]{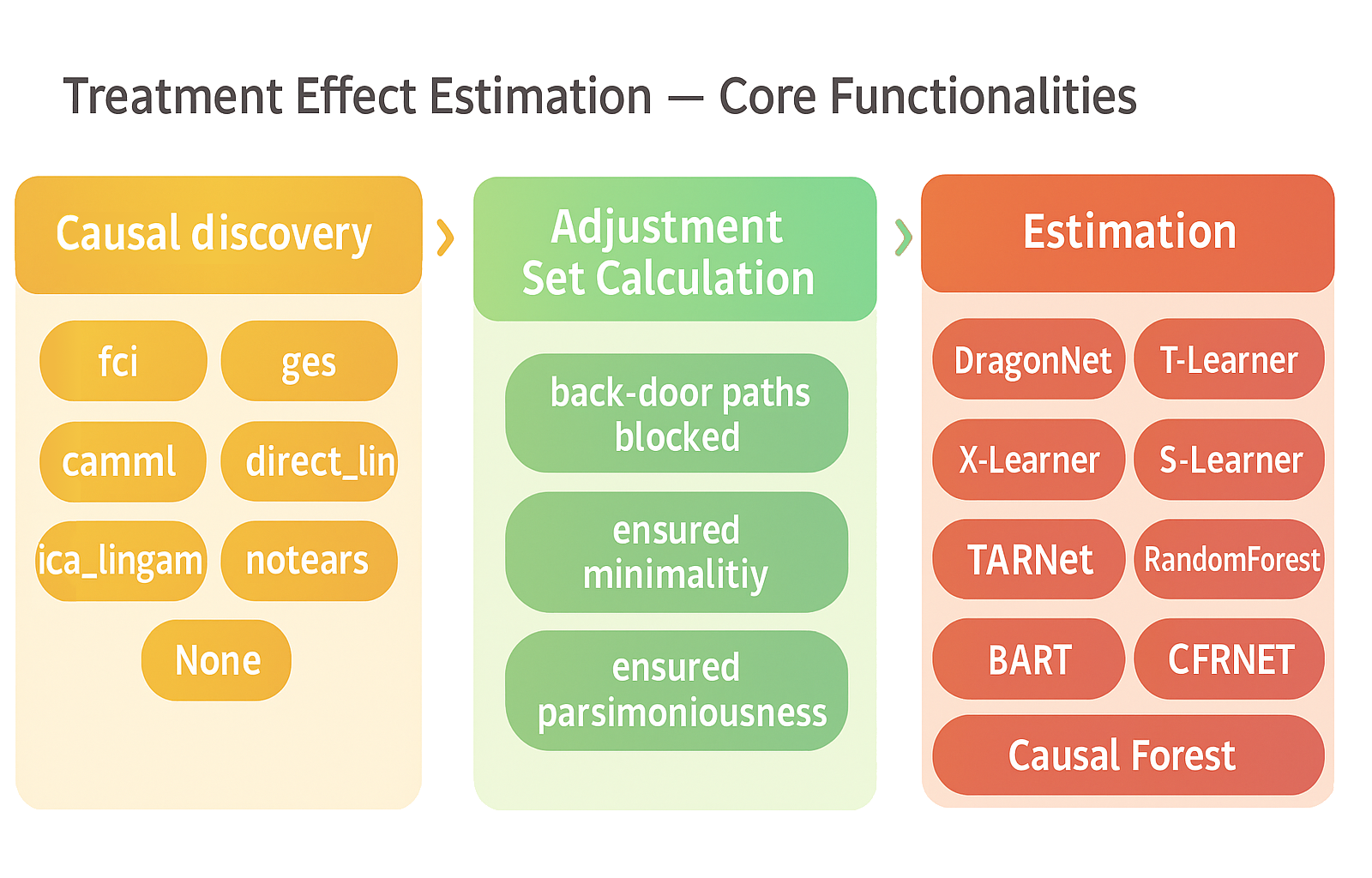}
    \caption{The core functions in CATE-B showing all the causal discovery, treatment effects estimation, and Adjustment set calculation methods}
    \label{fig:core-functions}
\end{figure}
Figure \ref{fig:core-functions} shows the plug-ins that are provided with the package under the three key steps for the treatment effects estimation pipeline. The high-level architecture of CATE-B, is organized into five modular components:

\textbf{Data Input}. Responsible for ingesting and validating user data (e.g.\ tabular CSV) and inferring schema triggered by a drag-and-drop upload widget. 

\textbf{DAG Discovery Layer}. Exposes a uniform and extensible DAG discovery interface to any graph-learning algorithm. Out-of-the-box we support PC, GES, CAMML, direct-lingam, ica-lingam, and NO-TEARS, but new methods can be added simply by adding providing additional plug-ins. Each plug-in returns either a single DAG or a Markov equivalence class, which the system forwards to the next stage.

\textbf{Effect Estimation Layer}. Provides a common and extensible estimate interface for any ATE estimator. Built-in estimators a suite of neural-network-based meta-learners (e.g.\ TARNET, CFRNET), but developers can drop in custom regressors by adding new Effect estimation plug-ins.

\textbf{LLM Coordinator \& Chatbot Front-end}. Acts as a “strategic planner” that interprets natural-language queries, decides which discovery or estimation plug-ins to invoke, and sequences do-calculus, adjustment-set derivation, and sensitivity checks into a coherent pipeline. The LLM engine itself is fully abstracted behind a lightweight adapter, so that different models (e.g.\ GPT-4, OpenChat 3.5) can be swapped in without changing the core logic.

\textbf{Results \& Reporting}. Collects intermediate artifacts (discovered graph, adjustment sets, propensity scores, effect estimates) into a centralized “state bank,” then renders visualizations and data tables. The pipeline concludes with a list of assumptions that have been made during the session. Explicitly listing the assumptions ensures complete transparency of a topic that is often opaque to the non-expert in causality, and thus makes sure that no incorrect assumptions are made. The user can also trackback to remove invalid assumptions in the pipeline to ensure the estimation is valid. 

Together, these components form an extensible, human-in-the-loop framework: a coordinator module operates over the state bank to plan each causal-analysis step, while a worker module executes plug-in calls and updates the state. Domain experts interact via the chatbot interface to provide feedback or override assumptions, and the system logs every decision to ensure full reproducibility. Thanks to the plug-in abstractions, new graph-discovery methods, estimators, or even entirely new causal-inference paradigms can be integrated with minimal boilerplate-simply implement the base interface, register your plug-in, and will incorporate it into its conversational workflows.

A full, in-dept, overview of these core functionalities is provided in our appendix. Our appendix also shows how one can customize \texttt{CATE-B} with additional plugins or through inheriting other class functionalities. 

\vspace{-10pt}
\subsection{Causal Pipeline}
Data flows through CATE-B according to the user’s chosen workflow:

\textbf{Implicit CATE Workflow.} Users who prefer a classical meta-learner can bypass DAG recovery entirely. After upload, they select a CATE learner (e.g. DR-Learner) from the Effect Estimation Layer and invoke estimate(data, implicit=True). The system treats all observed covariates as the adjustment set and proceeds directly to regression.

\textbf{SCM-Driven Workflow.} Users desiring explicit causal structure first recover an SCM; the chosen DAG plug-in returns a graph or MEC, which the platform then passes to an adjustment-set derivation routine. Once the minimal adjustment set is computed, users pick either a simple regressor (e.g. random forest) or a CATE learner.

\section{Experimental Results} \label{sec:experiments}
Table~\ref{tab:discovery-ate-grid} and Table \ref{tab:ate-by-discovery-survey} compare the estimated average treatment effect (ATE) of the CATE-B benchmarking environment.  We report the mean estimate over repeated runs and the corresponding empirical standard deviation. The values clearly show the importance of accurate causal discovery in the ATE pipeline. The results also highlight the importance of the Causal discovery method in finding the correct Adjustment set.

\begin{table}[htbp]
\vspace{-10pt}
  \centering
  \scriptsize
  \resizebox{\textwidth}{!}{%
    \begin{tabular}{lccccccc}
      \toprule
      & \multicolumn{7}{c}{\textbf{Causal Discovery Method}} \\
      \cmidrule(l){2-8}
      \textbf{ATE Method}         & \textbf{fci}       & \textbf{ges}        & \textbf{camml}      & \textbf{direct\_lingam} & \textbf{ica\_lingam} & \textbf{notears}     & \textbf{None}        \\
      \midrule
      \textbf{Adjustment set}     & Proteinuria + Age  & Age                & Age                & Age                    & Age                  & Age                  & All covariates                  \\
      \midrule
      DragonNet                   & $-0.554 \pm 2.364$ & $0.917 \pm 1.186$  & $1.294 \pm 1.222$  & $1.173 \pm 0.933$      & $1.132 \pm 1.005$    & $1.763 \pm 0.844$    & $-0.325 \pm 2.168$   \\
      T‐Learner                   & $-0.991 \pm 0.006$ & $0.974 \pm 0.008$  & $0.972 \pm 0.009$  & $0.980 \pm 0.007$      & $0.974 \pm 0.007$    & $0.975 \pm 0.008$    & $-0.974 \pm 0.073$   \\
      X‐Learner                   & $-0.913 \pm 0.005$ & $0.937 \pm 0.006$  & $0.936 \pm 0.009$  & $0.943 \pm 0.005$      & $0.938 \pm 0.007$    & $0.937 \pm 0.008$    & $-0.895 \pm 0.003$   \\
      TARNet                      & $-0.050 \pm 1.653$ & $1.132 \pm 1.231$  & $1.202 \pm 1.720$  & $1.673 \pm 1.632$      & $1.404 \pm 0.633$    & $1.175 \pm 1.597$    & $-0.779 \pm 3.163$   \\
      S‐Learner                   & $-0.357 \pm 0.010$ & $0.787 \pm 0.004$  & $0.787 \pm 0.008$  & $0.785 \pm 0.009$      & $0.786 \pm 0.004$    & $0.785 \pm 0.009$    & $-0.344 \pm 0.007$   \\
      Simple regression           & $-0.343 \pm 0.009$ & $0.782 \pm 0.004$  & $0.783 \pm 0.009$  & $0.780 \pm 0.009$      & $0.781 \pm 0.004$    & $0.781 \pm 0.010$    & $0.785 \pm 0.008$    \\
      BART                        & $-0.984 \pm 0.006$ & $0.959 \pm 0.006$  & $0.957 \pm 0.009$  & $0.965 \pm 0.006$      & $0.960 \pm 0.007$    & $0.959 \pm 0.008$    & $-0.972 \pm 0.007$   \\
      CFRNET                      & $-0.929 \pm 0.080$ & $1.256 \pm 0.037$  & $1.262 \pm 0.036$  & $1.212 \pm 0.029$      & $1.252 \pm 0.038$    & $1.218 \pm 0.051$    & $-0.977 \pm 0.008$   \\
      Causal Forest               & $-0.969 \pm 0.019$ & $0.971 \pm 0.013$  & $0.970 \pm 0.006$  & $0.967 \pm 0.006$      & $0.970 \pm 0.007$    & $0.967 \pm 0.007$    & $0.786 \pm 0.011$    \\
      \midrule
      Ground truth                & $1.050$ & $1.050$ & $1.050$ & $1.050$ & $1.050$ & $1.050$ & $1.050$ \\
      \bottomrule
    \end{tabular}
  }
  \caption{Average treatment effect (ATE) estimates (mean~$\pm$~std.\ dev.) for each combination of causal‐discovery method and ATE learner for the sodium dataset. The causal effect being estimated is the effect of sodium intake on blood pressure. The values presented are the means and standard deviations across 10 runs. The "None" value indicates that there is no causal discovery and the adjustment set that is used for the ATE is the superset of all covariates.}
  \label{tab:discovery-ate-grid}
\end{table}

\begin{table}[htbp]
  \centering
  \scriptsize
  \resizebox{\textwidth}{!}{%
    \begin{tabular}{lccccc}
      \toprule
      & \multicolumn{5}{c}{\textbf{Causal Discovery Method}} \\
      \cmidrule(l){2-6}
      \textbf{ATE Method}       & \textbf{FCI}           & \textbf{CAMML}        & \textbf{Direct LiNGAM}  & \textbf{ICA LiNGAM}     & \textbf{None}          \\
      \midrule
      \textbf{Adjustment set}   & Residence + Age        & Age                   & Residence + Age         & Age                     & All                    \\
      \midrule
      DragonNet                 & $-0.005 \pm 0.003$     & $-0.007 \pm 0.004$    & $-0.001 \pm 0.004$      & $-0.008 \pm 0.006$      & $0.006 \pm 0.006$      \\
      T‐Learner                 & $0.000 \pm 0.002$      & $-0.001 \pm 0.002$    & $0.001 \pm 0.002$       & $-0.001 \pm 0.002$      & $0.011 \pm 0.002$      \\
      X‐Learner                 & $0.000 \pm 0.002$      & $-0.001 \pm 0.002$    & $0.001 \pm 0.002$       & $-0.001 \pm 0.002$      & $0.012 \pm 0.002$      \\
      TARNet                    & $-0.003 \pm 0.004$     & $-0.008 \pm 0.003$    & $-0.003 \pm 0.004$      & $-0.007 \pm 0.003$      & $0.007 \pm 0.007$      \\
      S‐Learner                 & $0.001 \pm 0.002$      & $-0.008 \pm 0.003$    & $0.002 \pm 0.001$       & $0.000 \pm 0.002$       & $0.011 \pm 0.001$      \\
      Simple regression         & $0.001 \pm 0.002$      & $-0.001 \pm 0.001$    & $0.002 \pm 0.001$       & $0.000 \pm 0.002$       & $0.011 \pm 0.002$      \\
      BART                      & $0.000 \pm 0.002$      & $-0.001 \pm 0.002$    & $0.001 \pm 0.002$       & $-0.001 \pm 0.002$      & $0.012 \pm 0.002$      \\
      CFRNET                    & $0.000 \pm 0.004$      & $-0.001 \pm 0.005$    & $-0.001 \pm 0.005$      & $0.000 \pm 0.005$       & $0.004 \pm 0.007$      \\
      Causal Forest             & $0.001 \pm 0.003$      & $-0.001 \pm 0.002$    & $0.001 \pm 0.003$       & $0.000 \pm 0.002$       & $0.006 \pm 0.004$      \\
      \bottomrule
    \end{tabular}
  }
  \caption{Average treatment effect (ATE) estimates (mean~$\pm$~std.\ dev.) for each combination of causal‐discovery method and ATE learner for the survey dataset. The causal effect being measured here is the effect of sex on occupation. The values presented are the means and standard deviations across 10 runs. The "None" value indicates that there is no causal discovery and the adjustment set that is used for the ATE is the superset of all covariates.}
  \label{tab:ate-by-discovery-survey}
\end{table}

These results demonstrate that CATE-B combined structural discovery and estimation workflow achieves greater accuracy, while still keeping track of all assumptions. All of this is achieved without additional plugins our other code inferfacing with \texttt{CATE-B}, allowing users access to state-of-the-art causal inference techniques straight out of the box.

\section{Conclusion}

We introduce \texttt{CATE-B}, a novel co-pilot for treatment effect estimation that leverages large language models and causal discovery algorithms to bridge the persistent gap between causal inference theory and its practical deployment. By combining scientific literature retrieval, LLM-powered reasoning, and epistemic uncertainty quantification, our system facilitates the recovery of structural causal models from observational data—even in domains where causal expertise is limited or prohibitively expensive to access. Central to this workflow is the notion of the Minimum Uncertainty Adjustment Set (MUAS), which systematically minimizes reliance on fragile causal assumptions, thereby yielding more robust and trustworthy treatment effect estimates.

To support reproducibility and future research, we release a benchmark suite of real-world datasets, curated prompts, causal graph annotations, and automated evaluations designed specifically to test LLM-augmented causal inference pipelines. Our benchmark provides a first-of-its-kind infrastructure to evaluate causal discovery and adjustment strategies not only by accuracy but also by robustness to uncertainty and mis-specification.

We envision \texttt{CATE-B} as a stepping stone toward a new class of domain-aware, epistemically grounded AI systems—where scalable causal inference becomes accessible not just to statisticians, but to practitioners and decision-makers across disciplines. We hope this work catalyzes a broader conversation about how causal benchmarks can—and must—evolve to meet the needs of real-world deployment. As treatment effect estimates increasingly guide decisions in healthcare, education, and public policy, ensuring their transparency, robustness, and domain-relevance is not merely a technical concern, but a societal imperative.

\subsubsection*{Acknowledgments}
Use unnumbered third level headings for the acknowledgments. All
acknowledgments, including those to funding agencies, go at the end of the paper.
Only add this information once your submission is accepted and deanonymized. 

\bibliography{references}
\bibliographystyle{tmlr}

\newpage
\appendix
\section{Appendix}

\section{Limitations on the Datasets}
\paragraph{Dataset limitations.}
CATE-B is designed for settings with a binary treatment and sufficient signal for structure learning and statistical testing. Datasets are admissible provided they meet the constraints below; if not, cautious pre-processing (e.g., variable pruning, category consolidation, or sample augmentation) may mitigate—but not eliminate—associated risks.

\begin{itemize}
  \item \textbf{Binary treatment:} the dataset must include a designated treatment indicator \(T \in \{0,1\}\).
  \item \textbf{Limited categorical covariates:} to preserve statistical power and computational tractability, the dataset should contain only a small number of categorical variables, each with few levels; the proliferation of low-variance indicators (e.g., many binary dummies) can weaken tests and destabilize DAG discovery.
  \item \textbf{Narrow (low-dimensional) feature set:} the dataset should be ``narrow'', with \(number of samples \gg number of features\), so that structure learning yields stable and accurate DAGs.
  \item \textbf{Domain-grounded variables:} included features should be well represented in the scientific literature (e.g., common biomedical measurements) to support defensible causal assumptions and external validation.
\end{itemize}

\subsection{plug-in Framework}
At the core of CATE-B’s extensibility is a simple, light-weight plug-in interface for both DAG discovery and ATE estimation. This modular approach not only allows user's to easily contribute their own methods, but also allows for maximum inter-operability of causal discovery and ATE estimation. On startup, the system automatically scans the plug-in directories, imports every module exposing that exposes the correct attributes/methods, and registers them under their \texttt{NAME}. During execution, the worker agent invokes \texttt{run} method for then collects the output, which is either the causal graph or the ATE.

\subsection{Dag Discovery}
Each plug-in lives in its own module, and inherits from a \texttt{DAGDiscoveryBase} class, see listing \ref{lst:dag-discovery-base}. This mandates the presence of a name property and a \texttt{\_run()} method in the plug-ins that inherit from it. The \texttt{\_run()} method takes in the data and outputs a signed adjacency matrix for either a concrete DAG or a Markov equivalence class. On startup, the system automatically scans the plug-in directory, imports every module exposing these two elements, and registers them under their NAME. During execution, the worker agent can invoke \texttt{run(data)} for any subset of registered plug-ins-either individually or in parallel-and collect their graph outputs.

\begin{lstlisting}[caption={Base class for DAG discovery plug-ins}, label={lst:dag-discovery-base}]
class DAGDiscoveryBase:
    """
    Base class for DAG discovery plug-ins.

    Subclasses must implement the `NAME` property and the `_run` method,
    which returns an adjacency matrix (`np.ndarray` of shape [n,n])
    using the convention:
      - `-1` at [i,j] and `1` at [j,i] for directed i->j
      - `1` at both [i,j] and [j,i] for undirected i--j
      - `0` for absent edges.
    The base `run` method wraps the signed matrix into a `_SimpleGraph`.
    """
    def __init_subclass__(cls, **kwargs):
        super().__init_subclass__(**kwargs)
        instance = cls()
        if not isinstance(instance.NAME, str):
            raise TypeError(f"Subclass {cls.__name__!r} must define a NAME property returning a string")

    @property
    def NAME(self) -> str:
        raise NotImplementedError(f"{self.__class__.__name__} must implement the NAME property")

    @classmethod
    def run(cls, data: np.ndarray, node_names: List[str], **kwargs: Any) -> dict:
        # Delegate to subclass implementation to get signed adjacency matrix
        signed = cls._run(data, node_names, **kwargs)
        # Wrap into a SimpleGraph and return
        graph_obj = _SimpleGraph(signed, node_names)
        return {"G": graph_obj}

    @classmethod
    def _run(cls, data: np.ndarray, node_names: List[str], **kwargs: Any) -> np.ndarray:
        """
        Subclasses must implement this method to compute the signed adjacency matrix.
        """
        raise NotImplementedError(f"{cls.__name__} must implement the _run() method")
\end{lstlisting}

\subsection{ATE estimation}

Each plug-in lives in its own module, and inherits from a \texttt{ATEEstimationBase} class, see listing \ref{lst:ATE-estimation-base}. This forces the ATE estimation plug-ins to instantiate a name property and a \texttt{\_estimate\_effect\_once()} method in the plug-ins that inherit from it. The \texttt{\_estimate\_effect\_once()} method takes in the covariate matrix, as well as the treatment and outcome variables. The downstream plug-in must output an estimation of the causal effect. 

\begin{lstlisting}[caption={Base class for ATE plug-ins}, label={lst:ATE-estimation-base}]
class ATEEstimationBase:
    """
    Base class for ATE estimation plug-ins. Handles looping over seeds,
    computing ATE and (optional) PEHE, and summarizing results.

    Subclasses must implement:
      - a @property NAME: str
      - a classmethod _estimate_effect_once(Y, T, X, seed, **kwargs) -> np.ndarray
    """

    def __init_subclass__(cls, **kwargs):
        super().__init_subclass__(**kwargs)
        instance = cls()
        if not isinstance(instance.NAME, str):
            raise TypeError(
                f"Subclass {cls.__name__!r} must define a NAME property returning a string"
            )

    @property
    def NAME(self) -> str:
        raise NotImplementedError("Subclasses must override the NAME property")

    @classmethod
    def run(
        cls,
        Y: np.ndarray,
        T: np.ndarray,
        X: np.ndarray,
        n_runs: int = 10,
        tau_true: Optional[np.ndarray] = None,
        **kwargs: Any
    ) -> Dict[str, float]:
        """
        Orchestrates n_runs repetitions of effect estimation, aggregates ATE and PEHE.

        Parameters
        ----------
        Y : array-like
            Outcomes.
        T : array-like
            Binary treatment indicators.
        X : array-like
            Covariates.
        n_runs : int
            Number of bootstrap runs / random seeds.
        tau_true : array-like, optional
            True individual treatment effects; if provided, PEHE is computed.

        Returns
        -------
        Dict[str, float]
            'ate_mean', 'ate_std', 'pehe_mean', 'pehe_std'
        """
        ate_list: List[float] = []
        pehe_list: List[float] = []

        for seed in range(n_runs):
            tau_hat = cls._estimate_effect_once(
                Y, T, X, seed=seed, **kwargs
            )
            ate_list.append(float(np.mean(tau_hat)))

            if tau_true is not None:
                pehe_list.append(
                    float(np.sqrt(np.mean((tau_hat - tau_true) ** 2)))
                )
            else:
                pehe_list.append(float('nan'))

        return {
            "ate_mean": round(float(np.mean(ate_list)), 4),
            "ate_std":  round(float(np.std(ate_list)), 4),
            "pehe_mean": round(float(np.nanmean(pehe_list)), 4),
            "pehe_std":  round(float(np.nanstd(pehe_list)), 4),
        }

    @classmethod
    def _estimate_effect_once(
        cls,
        Y: np.ndarray,
        T: np.ndarray,
        X: np.ndarray,
        seed: int,
        **kwargs: Any
    ) -> np.ndarray:
        """
        Run one seed's worth of training and effect estimation.

        Must return tau_hat: an array of individual treatment effects.
        """
        raise NotImplementedError(
            f"{cls.__name__} must implement _estimate_effect_once()"
        )
\end{lstlisting}

\subsection{LLM Coordinator Layer}
The LLM Coordinator serves as the “strategic planner” of the system, sequencing causal-analysis steps under the hood. It communicates with the conversational front end via dynamic high-level plan stored in JSON formatted tasks with descriptions. Different plan steps have access to different functions and invokes the corresponding plug-in calls in the correct order. All interaction with the language model-prompt templates and answer parsing-passes through an engine interface, which encapsulates model-specific details such as API endpoints or rate limits. This abstraction allows one to swap backend models (GPT-4, OpenChat 3.5 or potentially non-function calling LLMs like Llama) by providing an alternative adapter, without touching the coordinator logic. The Coordinator maintains a structured “plan” object-modeled after multi‐agent reasoning architectures-that it incrementally refines based on plug-in outputs and user feedback, ensuring each next step is both contextually appropriate and responsive to domain guidance.

\subsection{Chatbot Interface}
The chatbot front-end is the user’s entry point into CATE-B’s workflow. It implements a standard natural‐language understanding pipeline, backed by a simple state machine that tracks the current causal‐analysis phase and holds context (e.g. uploaded schema, discovered graph, selected adjustment set). When ambiguity arises, such as multiple candidate adjustment sets, the interface engages in clarification loops, asking targeted questions, e.g. “Which adjustment set would you like to use (given this description of what each means ...)?”. All interactions, decisions, and system outputs are logged in the state bank to support auditability and reproducibility.

\end{document}

%% file: math_commands.tex
\usepackage{amsmath,amsfonts,bm}

\def\eqref#1{equation~\ref{#1}}
\def\1{\bm{1}}

\DeclareMathAlphabet{\mathsfit}{\encodingdefault}{\sfdefault}{m}{sl}
\SetMathAlphabet{\mathsfit}{bold}{\encodingdefault}{\sfdefault}{bx}{n}